\documentclass[conference]{IEEEtran}
\IEEEoverridecommandlockouts
\usepackage{amsmath,amssymb,amsfonts}
\usepackage{algorithmic}
\usepackage{textcomp}
\usepackage{xcolor}
\def\BibTeX{{\rm B\kern-.05em{\sc i\kern-.025em b}\kern-.08em
    T\kern-.1667em\lower.7ex\hbox{E}\kern-.125emX}}

\usepackage[numbers]{natbib}
\usepackage{booktabs}
\usepackage{multirow}
\usepackage{graphicx}
\usepackage{listings}
\usepackage{relsize}
\usepackage[color=pink]{todonotes}
\usepackage{hyperref}
\usepackage[capitalize,noabbrev]{cleveref}
\usepackage{comment}
\usepackage{subcaption}

\newcommand{\STAB}[1]{\begin{tabular}{@{}c@{}}#1\end{tabular}}
    
\usepackage{CJKutf8}

\begin{document}

\title{A Neural Approach for Detecting Morphological Analogies
%Conference Paper Title%*\\
}

%\begin{comment}
\author{\IEEEauthorblockN{Safa Alsaidi$^*$}
\IEEEauthorblockA{\textit{IDMC}, % \\
\textit{Université de Lorraine}\\
Nancy, France \\
safaa-98@hotmail.com}

\and
\IEEEauthorblockN{Amandine Decker$^*$}
\IEEEauthorblockA{\textit{IDMC}, % \\
\textit{Université de Lorraine}\\
Nancy, France \\
decker.amandine@hotmail.com}

\and
\IEEEauthorblockN{Puthineath Lay$^*$}
\IEEEauthorblockA{\textit{IDMC}, % \\
\textit{Université de Lorraine}\\
Nancy, France \\
puthineathlay@gmail.com}

\and
\IEEEauthorblockN{Esteban Marquer}
\IEEEauthorblockA{\textit{Université de Lorraine, CNRS, LORIA} \\
F-54000, France \\
esteban.marquer@loria.fr}
\and
\IEEEauthorblockN{Pierre-Alexandre Murena}
\IEEEauthorblockA{
%\textit{Helsinki Institute for Information Technology HIIT} \\
\textit{%Department of Computer Science
HIIT, Aalto University}\\
Helsinki, Finland \\
pierre-alexandre.murena@aalto.fi}
\and
\IEEEauthorblockN{Miguel Couceiro}
\IEEEauthorblockA{\textit{Université de Lorraine, CNRS, LORIA} \\
F-54000, France \\
miguel.couceiro@loria.fr}
}
%\end{comment}
%\begin{comment}
%\author{\IEEEauthorblockN{Anonymous}
%\IEEEauthorblockA{Anonymous }
%}
%\end{comment}

\maketitle

\begingroup\renewcommand\thefootnote{*}
\footnotetext{Authors contributed equally.}
\endgroup

\begin{abstract}
Analogical proportions are statements of the form ``A is to B as C is to D'' that are used for several reasoning and classification tasks in artificial intelligence and natural language processing (NLP).
For instance, there are analogy based approaches to semantics as well as to morphology.
In fact, symbolic approaches were developed to solve or to detect analogies between character strings, e.g., the axiomatic approach  as well as that based on Kolmogorov complexity. 
In this paper, we propose a deep learning approach to detect morphological analogies, for instance, with reinflexion or conjugation.
We present empirical results that show that our framework is competitive with the above-mentioned state of the art symbolic approaches.
We also explore empirically its transferability capacity across languages, which highlights interesting similarities between them.
%Our result is state of the art performance
%In this paper, we adapt the deep learning approach that  developed for semantical anaogies to morphology and achieve state of the art performance, competve wrt. symbolic aproaches
\end{abstract}% no math or spetial sym


\section{Introduction}
An analogy, or analogical proportion, is a relation between four elements $A$, $B$, $C$, and $D$ meaning ``$A$ is to $B$ as $C$ is to $D$'', often written as $A:B::C:D$.
A typical example would be the semantic analogy ``\textit{kitten} is to \textit{cat} as \textit{puppy} is to \textit{dog}''.

Analogy is a major component of human cognition~\cite{analogy-making-cognition:2001:mitchel,analogy-making-ai:2021:mitchel} and of the process of learning by extrapolation~\cite{extend-boolean:2017:couceiro}, and multiple cognitive linguistics theories support the importance of analogies in learning and using a language~\cite{linguistique-generale:1916:saussure,bod2009exemplar,behrens2017role}.
%For example, in Construction Grammar semantic is attached to \textit{constructions}, patterns modified by analogy to fit the need of the speaker.\todo{check veracity, then cite or discard}
Analogies have been extensively studied in NLP, resulting in different formalizations (like the one described in \cref{sec:formal}) with noteworthy applications such as derivational morphology \cite{tools-analogy:2018:fam-lepage,minimal-complexity:2020:murena} and machine translation \cite{nn-analogy-translation-fr:2020:taillandier}.

Analogical reasoning can capture different aspects of the manipulated objects. For example, analogies on words can refer exclusively to their morphology (``\textit{cats} is to \textit{cat} as \textit{trees} is to \textit{tree}'') or their  semantics (``\textit{kitten} is to \textit{puppy} as \textit{cat} is to \textit{dog}'').
The question of the correctness of an analogy $A:B::C:D$ is a difficult task and it has been tackled using both  formal  \cite{tools-analogy:2018:fam-lepage,trends-analogical-reasoning:2014:prade} and empirical approaches \cite{minimal-complexity:2020:murena,analogies-ml-perspective:2019:lim,3-cos-mul:2014:levy-goldberg,efficient-representation:2013:mikolov,eval-vector-analogy:2017:chen}. Although the original challenge is to find some structure from $A$, $B$, $C$, and $D$, recent empirical works propose  data-oriented strategies based on machine learning to learn the correctness of analogies from past observations.
In particular, Lim \textit{et al.}~\cite{analogies-ml-perspective:2019:lim} propose   a deep learning approach  to train models of analogies on corpora of semantic analogies and pretrained GloVe embeddings \cite{glove:2014:pennington}. This approach achieves competitive results on analogy classification and completion.

In this paper, we adapt the approach developed by Lim \textit{et al.}~\cite{analogies-ml-perspective:2019:lim} for semantic word analogies to morphological analogies.
The major difference between the two approaches is that ours relies on a morphological word embedding, which had to be entirely developed and trained. Indeed, the pretrained embeddings used in \cite{analogies-ml-perspective:2019:lim} are adapted to semantic analogies only, and there are currently no readily available morphological word embeddings. 
%We differ from this approach as we focus on morphological analogies, and thus rely on morphological word embeddings. Moreover, we could not rely on pretrained embeddings like the ones used in \cite{analogies-ml-perspective:2019:lim} and we  had to develop and train our own, since,  opposed to semantic word embeddings, there are no readily available morphological word embeddings.
Furthermore, we explore the potential of transferring our neural analogy model across languages.

Our method achieves competitive results to state of the art symbolic approaches, and obtains promising results in  transferability settings. This shows the potential of Lim \textit{et al.}'s approach \cite{analogies-ml-perspective:2019:lim} for applications other than semantic analogies. We illustrate this fact with morphological analogies, and that opens the possibility of tackling multimodal analogies.
As long as we have a fitting embedding model, this approach could be further extended to analogies between domains that are usually considered too complex, such as analogies between functions or other complex models.

This paper is organized as follows. We recall in \cref{sec:related} a simple formalization of analogy and present related works, in particular, the approach of Lim \textit{et al.} \cite{analogies-ml-perspective:2019:lim} by which our framework is inspired.
In \cref{sec:approach} we explain how we adapt this approach to morphological analogy.
An empirical study carried out on the Sigmorphon2016~\cite{sigmorphon:2016:cotterell} and Japanese Bigger Analogy~\cite{jap-data:2018:karpinska} datasets is presented in \cref{sec:expe}, in which we compare our results to other state of the art approaches to morphological analogy, that show the competitiveness of our approach.
We also explore how the models trained on each language perform when transferred to the other languages in \cref{sec:transfer}, that reveal interesting similarities between them.

\section{Related Works}\label{sec:related}
%\todo[inline]{add intro §}
In this section we briefly discuss the axiomatic setting of analogy and address two main problems dealing with analogical reasoning, namely analogical identification (or classification) and inference (or regression) (\cref{sec:formal}). We then discuss different approaches for modeling analogies on words, from a morphological (\cref{sec:approaches-morpho}) and semantic (\cref{sec:approaches-sem}) point of view.

\subsection{Formal Analogy}\label{sec:formal}

Multiple attempts have been made to formalize and manipulate analogies, starting as early as De Saussure's work in 1916 \cite{linguistique-generale:1916:saussure}. We refer to \cite{analogy-formal:2004:lepage} for an introduction to analogy in linguistics. % in English and to \cite{analogy-commutation-linguistic-fr:2003:lepage} for a comprehensive introduction to analogy in linguistics in French.
 Most of the recent works on analogy use a formalization based on that of \cite{analogy-commutation-linguistic-fr:2003:lepage}, which is related to the  common view of analogy as a geometric (\cref{eq:geom-proportion}) or arithmetic proportion (\cref{eq:arith-proportion}) 
or, in geometrical terms, as a parallelogram in a vector space (\cref{eq:parallelogram}).
 \begin{equation}
  \frac{A}{B}=\frac{C}{D}
  \label{eq:geom-proportion}
 \end{equation}
 \begin{equation}
  A-B=C-D
  \label{eq:arith-proportion}
 \end{equation}
\begin{equation}
    \vec{A}-\vec{B}=\vec{C}-\vec{D}
    \label{eq:parallelogram}
\end{equation}
This formalism takes an axiomatic approach to define  \emph{proportional analogies}~\cite{trends-analogical-reasoning:2014:prade,analogical-dissimilarity:2009:miclet}: four objects $A$, $B$, $C$, and $D$ are in \emph{analogical proportion} (\textit{i.e.} $A:B::C:D$) if and only if the following three postulates hold true:
%This formalism is based on three main postulates:
\begin{itemize}
    \item $A:B::A:B$ (reflexivity);
    \item $A:B::C:D\rightarrow C:D::A:B$ (symmetry);
    \item $A:B::C:D\rightarrow A:C::B:D$ (central permutation).
\end{itemize}
These postulates imply several other properties: 
\begin{itemize}
    \item $A:A::B:B$ (identity); 
      \item $A:B::C:D\rightarrow B:A::D:C$ (inside pair reversing); 
        \item $A:B::C:D\rightarrow D:B::C:A$ (extreme permutation).
\end{itemize}
It can be easily verified that the geometric and arithmetic proportions (\cref{eq:arith-proportion,eq:geom-proportion}) and the parallelogram rule (\cref{eq:parallelogram}) define proper analogies.
Another frequently accepted postulate is uniqueness: for a triple $\langle A,B,C\rangle$, if there exists $D$ such that $A:B::C:D$ then $D$ is unique, which entails that $A:A::B:C \rightarrow C=B$.
Most of the works on analogy relying on such a formalization concern analogy between symbols and strings of symbols ($abc:abd::efg:efh$), which can be directly related to morphology ($cat:cats::dog:dogs$).
An example out of this scope is \cite{boolean-analogy:2018:couceiro} which deals with boolean functions.

Analogy related tasks can be divided in two main axes: \begin{itemize}
    \item {\it analogy identification}, in which given a quadruple $\langle A,B,C,D\rangle$ we want to know whether $A:B::C:D$ is a valid analogy \cite{tools-analogy:2018:fam-lepage,analogies-ml-perspective:2019:lim}; this can be extrapolated to analogy extraction from sets of objects  as in \cite{tools-analogy:2018:fam-lepage};
    \item {\it analogical inference}, which aims at solving ``analogical equations'', \textit{i.e.}, finding  the indeterminate $x$ in  expressions of the form   $A:B::C:x$  \cite{minimal-complexity:2020:murena,analogies-ml-perspective:2019:lim,analogy-alea:2009:langlais}.
\end{itemize}
When the manipulated objects have a vector representation (typically with embeddings), analogy identification can be seen as a binary classification task, whereas analogy solving can be seen as a regression task.
In this paper, we focus on analogical classification that can have a wide variety of linguistic applications as illustrated in \cite{tools-analogy:2018:fam-lepage}.

\subsection{Approaches to Model Analogies Between Symbols and Morphological Analogies}\label{sec:approaches-morpho}

Most algorithmic approaches to solving morphological analogies  rely on the aforementioned formal characterization of proportional analogies. Recent approaches include those by Fam and Lepage (2018)~\cite{tools-analogy:2018:fam-lepage} and the \lstinline{Alea} algorithm proposed by Langlais \textit{et al.} (2009)~\cite{analogy-alea:2009:langlais}.
Fam and Lepage's approach relies on feature vectors to detect analogies within a list of words and create analogical grids, allowing to build analogies between more than 4 words.
The axioms of Lepage \cite{analogy-formal:2004:lepage} have been also reformulated by Yvon \cite{finite-state-trancducers:2003:yvon} to give a closed form solution, which is computed by \lstinline{Alea} in a Monte-Carlo setting. In practice, \lstinline{Alea} uses random slicing and merging of the character strings $A$, $B$ and $C$ to obtain potential solutions to $A:B::C:x$, which are then ranked by how frequently they appear when repeating the random process a certain number of times. Good results can be obtained with 1000 receptions~\cite{analogy-alea:2009:langlais}.

A more empirical approach was proposed by Murena \textit{et al.} (2020)~\cite{minimal-complexity:2020:murena},  which relaxes the formal definition of analogical proportion. Following preliminary evidences that humans may follow a simplicity principle when solving analogies~\cite{chater2003simplicity,cornuejols1998analogy,murena2017complexity}, the authors propose to solve analogical equations $A:B::C:x$ by finding the $x$ that minimizes the total description length (or Kolmogorov complexity) of $A:B::C:x$. The total description length is evaluated using a simple description language for character strings and an associated binary code. 
Both \lstinline{Alea} and Fam and Lepage's approaches were shown to be outperformed by Murena \textit{et al.}'s Kolmogorov complexity approach on analogies, on the Sigmorphon2016 dataset.

\subsection{Approaches to Model Semantic Analogies}\label{sec:approaches-sem}
There is a long tradition of works on word representation using statistical methods based on co-occurrence of words to represent them as vectors, that we usually call \textit{word embeddings}.
Since early works on embedding spaces \cite{efficient-representation:2013:mikolov}, it is usually assumed that embeddings represent the semantics of words and that analogies in the vector space can be defined by simple arithmetic operations, either by relying on the parallelogram rule~\cite{eval-vector-analogy:2017:chen} (see \cref{eq:parallelogram-}) or by using a cosine distance~\cite{eval-vector-analogy:2017:chen} such as \cref{eq:cossine}
or \textit{3CosMul} \cite{3-cos-mul:2014:levy-goldberg}; other similar metrics can be found in  \cite{embedding-analogies:2016:drozd}.
\begin{equation}\label{eq:parallelogram-}
    \vec{B}-\vec{A} + \vec{C}=\vec{D}
\end{equation}
\begin{equation}\label{eq:cossine}
    \frac{(\vec{A}-\vec{B})\cdot (\vec{C}-\vec{D})}{||\vec{A}-\vec{B}||\times ||\vec{C}-\vec{D}||}
\end{equation} 
For instance, GloVe or \textit{word2vec}~\cite{efficient-representation:2013:mikolov} embeddings are usually able to solve simple analogical equations using the parallelogram rule such as $\vec{cat}-\vec{kitten}+\vec{dog}\approx\vec{puppy}$.% is a typical example using the parallelogram rule and works well for example with GloVe or \textit{word2vec}\cite{efficient-representation:2013:mikolov} embeddings.%\footnote{Some marginal results can be observed for morphology as well, but these approaches are not adapted to morphological analogies as they are centered on representations of the semantics of words. Our preliminary experiments on GloVe and Bert confirmed the poor performance of such embedding methods for more extensive morphological analogy, so we won't compare to them further in this article.}
%\todo{While some morphological results can be observed as well, these approaches are not adapted to this problem, that's why we will not compare to them.}

%\subsection{Limitation of Symbolic Approaches to Model Semantic Analogies}
However, as mentioned in \cite{eval-vector-analogy:2017:chen}, those simplistic rules for analogy do not match human performance in all cases.
Additionally, some strict properties of formal analogy can contradict human intuition, like \emph{uniqueness} which is problematic when dealing with (quasi-)synonyms, which should be interchangeable solutions of analogical equations on word semantics.
Hence, it becomes necessary to relax the model of analogy to match human performance on analogies. %\todo{Limitation of Symbolic Approaches , as PA mentioned, a new section or to 2.1 ?}
While Murena \textit{et al.}~\cite{minimal-complexity:2020:murena} follow a simplicity principle on the transformation between words, Lim \textit{et al.} \cite{analogies-ml-perspective:2019:lim} propose to learn an analogy operator directly from analogies as an elegant way to get out of the simplistic models of parallelogram and cosine analogy. They use simple neural networks to learn analogies from a dataset of semantic analogies, and propose different models for classification and regression purposes. Moreover, they rely on the properties of formal analogies mentioned in \cref{sec:formal} to increase the amount of training data (as described in \cref{sec:data-augment}) and to ensure that models fit those properties to a certain degree.

\section{Approach}\label{sec:approach}
We adapt the approach developed by Lim \textit{et al.} \cite{analogies-ml-perspective:2019:lim} for semantic analogies to apply it to morphological analogies. Even though we use the same architecture for analogy classification (see \cref{sec:cnn}) and a similar training process (\cref{sec:data-augment}), 
our approach differs in the use of a custom embedding model (\cref{sec:emb}, based on \cite{morpho-thesis:2020:vania}) trained along the classifier.

\subsection{Simple Neural Model for Classification}\label{sec:cnn}
The architecture proposed in \cite{analogies-ml-perspective:2019:lim} for classifying analogies is a Convolutional Neural Network (CNN) taking as input the embeddings of size $n$ of four elements $A$, $B$, $C$, and $D$ stacked into a $n\times4$ matrix. The CNN has 3 layers as follows (see \cref{fig:char-embedding}).
\begin{itemize}
    \item A first convolutional layer with filter of $1\times2$ is applied on the embeddings, such that for each component $\cdot_i$ of the embedding vector, each filter spans across $A_i$ and $B_i$ simultaneously, and  across $C_i$ and $D_i$ simultaneously, with no overlap. In this architecture,  128 such filters are used, with a Regularized Linear Unit (ReLU) as activation function. This layer can be seen as extracting the relations $A:B$ and $C:D$.
    \item A second convolutional layer with 64 filters of $2\times2$ is applied on the resulting $128\times n\times 2$ matrix, after which the result is flattened into a $64  (n-1)$ unidimensional vector. This layer can be seen as comparing $A:B$ and $C:D$, and acts as the relation ``as'' in ``$A$ is to $B$ as $C$ is to $D$''. Similarly to the previous layer, the activation function is ReLU.
    \item In the third layer, the output of the second one is fed to a fully-connected layer (or feed-forward layer or perceptron) with a single output that we bind between 1 and 0 using a sigmoid activation.
\end{itemize}

\subsection{Embedding Model}\label{sec:emb}
The model from \cite{analogies-ml-perspective:2019:lim} relies on an extensively pretrained model to produce a vector representation of each word, called its {\it word embedding}. The analogy model takes the embeddings as inputs, and is thus dependent on the quality and information captured by the embeddings.

In their article, Lim \textit{et al.}~\cite{analogies-ml-perspective:2019:lim} work with semantic analogies on English words, and use a pretrained GloVe embeddings that are supposed to encode the semantics of  words with high fidelity.
However, we work on morphological analogies and on languages which do not necessarily have large embedding models readily available (see \cref{sec:data}), as most embedding models are trained on English.
In addition to the difficulty of obtaining embedding models for the languages we work on, we expected semantic embeddings to perform very poorly in a purely morphological task,  even if for languages like Chinese where morphology is strongly related to word semantic, that effect might be less important.
%To confirm our hypothesis we ran preliminary experiments using a German GloVe model\footnote{\url{https://int-emb-glove-de-wiki.s3.eu-central-1.amazonaws.com/vectors.txt}, from \url{https://deepset.ai/german-word-embeddings}}, resulting in very poor performance.\todo[inline]{get results with glove and the classification task}\todo[inline]{from PA: ``Here for instance, it would be fine to postpone the results to the experimental sections, depending on the size. If they are too long, here, they would interrupt the reading.''}

For example for German, one of the most readily available languages of the Sigmorphon2016 dataset, we initially experimented with a pretrained German GloVe model\footnote{\url{https://int-emb-glove-de-wiki.s3.eu-central-1.amazonaws.com/vectors.txt}, from \url{https://deepset.ai/german-word-embeddings}}.
However, this rather large GloVe model only has a coverage of 51.26\% (8872 of 17308 words) of the words present in the training set and 49.10\% (9798 of 19955 words) for the test set. This means a large portion of the analogies have at least one word unknown by the model, which is thus replaced by an embedding full of 0. Similar issues happened with other pretrained embeddings we tried.
Our other preliminary experiments using GloVe and Bert Multilingual~\cite{bert-multilingual:2018:delvin} confirmed the poor performance of such embedding methods for extensive morphological analogy even if some marginal results can be observed. Due to these issues we will not compare to them further in this article.

\begin{figure*}[htbp]
%\centerline{\includegraphics[width=.5\textwidth]{morpho_embed.png}}
\centerline{\includegraphics[width=\textwidth]{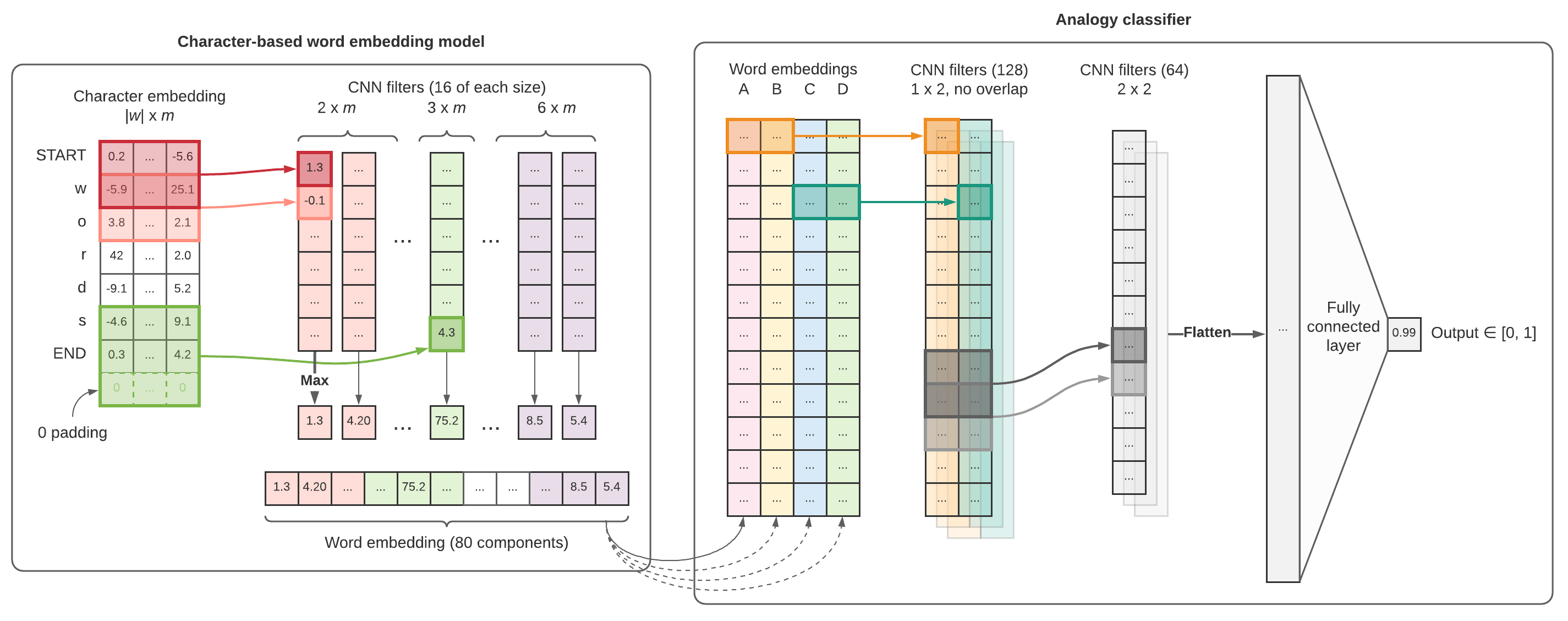}}
\caption{Overall structure of the CNN character-level word embedding model and CNN classifier.}
\label{fig:char-embedding}
\end{figure*}

To compensate for this issue, we use a character-level word embedding model able to capture the morphological aspect of words.
To our knowledge no such embedding model is available for the languages we manipulate, so we decided to train an embedding model together with our classifier.
We use a very simple model architecture based on Convolutional Neural Network (CNN) as described in \cite{morpho-thesis:2020:vania}.
This architecture is designed to embed individual words for morphological tasks.
It is composed of two layers as follows, also presented in \cref{fig:char-embedding}.
\begin{itemize}
    \item First we have a character embedding layer, which provides a learnable vector representation of size $m$ for each character. If a character is not seen during training, it's embedding is full of 0 by default. Additionally, we add a special ``start of word'' and an ``end of word'' character at each end of the word. For a word of  $|w|$ characters, we obtain a $(|w|+2)\times m$ matrix.
    \item Then, we have a convolutional layers with 16 filters of each of the sizes in $2\times m, 3\times m, \cdots, 6\times m$, for a total of $16\times5=80$ filters.
    The filters are used on the embeddings such that they overlap along the character dimension.
    Whenever necessary, 0 padding is used such that the center of each filter reach the extremities of the word, as shown in \cref{fig:char-embedding}.
    We expect those filters to capture morphemes\footnote{Morphemes are minimal morphological units which in our setup can be seen as n-grams.} of up to 6 characters.
    \item In the output layer, we use a max-pooling method to reduce the output of the convolutional layer along the character dimension, which gives us a vector of 80 components (one component per filter) used as the embedding.
\end{itemize}%\todo{Describe new embedding architecture}
%https://towardsdatascience.com/the-definitive-guide-to-bidaf-part-2-word-embedding-character-embedding-and-contextual-c151fc4f05bb
%https://claravania.github.io/assets/clara_phd_thesis.pdf

\subsection{Training Process}\label{sec:data-augment}
The model was trained for binary classification using \textit{binary cross entropy}, where the positive class is analogies and the negative is non-analogies.
Since the Sigmorphon2016 dataset we use in our experiments contains only positive examples (\textit{i.e.} correct analogical quadruples), we use the same data augmentation process as in \cite{analogies-ml-perspective:2019:lim} to generate negative examples (\textit{i.e.} invalid analogical quadruples).

This process was designed based on the properties of formal analogies, resulting in 8 equivalent forms for every analogy $A:B::C:D$:\begin{itemize}
    \item $A:B::C:D$ (base form);
    \item $C:D::A:B$ (symmetry);
    \item $A:C::B:D$ (central permutation);
    \item $B:A::D:C$ (inside pair reversing);
    \item $D:B::C:A$ (extreme permutation);
    \item $D:C::B:A$  (inside pair reversing followed by symmetry);
    \item $C:A::D:B$ (central permutation followed by inside pair reversing);
    \item $B:D::A:C$ (extreme permutation followed by inside pair reversing).
\end{itemize}
In addition to those equivalent forms, we have 3 analogical forms per valid analogy which are considered invalid analogies as they cannot be deduced from the base form $A:B::C:D$ without contradicting at least one of the postulates mentioned in \cref{sec:formal}:\begin{itemize}
    \item $B:A::C:D$;
    \item $C:B::A:D$;
    \item $A:A::C:D$.
\end{itemize}

Using this process we obtain a total of 8 positive and $3\times 8=24$ negative examples per analogy in the dataset, as each negative form has 8 equivalent forms.

%To train the embedding and classification model together for each analogy $A:B::C:D$, we first compute the embedding of $A$, $B$, $C$, and $D$ and train the classifier for all the permutations mentioned above. This ensures stability of the training process as, for example, it avoids that the embedding model to specialize for negative examples as they are 3 times more frequent in the training data.
To train the embedding and classification model together for each analogy $A:B::C:D$, we first compute the embedding of $A$, $B$, $C$, and $D$ and train the classifier for all the permutations mentioned above. 
Additionally, during training we do not apply the negative forms on each permutation of the analogy but only on the base form, resulting in 8 positive examples for 3 negative ones (instead of 8 and 24).
This ensures stability of the training process as, for example, it prevents the embedding model from specializing on negative examples as those would be 3 times more frequent in the training data. %\todo{if we retrain the embedding model, revert back that part}

\section{Experiments\label{sec:expe}}
In this section we detail our experimental setup on the Japanese Bigger Analogy Test Set~\cite{jap-data:2018:karpinska} and the Sigmorphon2016 dataset~\cite{sigmorphon:2016:cotterell} (both detailed in \cref{sec:data}). We also present how we used the approaches from \cite{minimal-complexity:2020:murena,analogy-alea:2009:langlais} as baselines (\cref{sec:baselines}).

The code used for our experiments is written in Python 3.9 and PyTorch and is available in the repository \url{https://github.com/AmandineDecker/nn-morpho-analogy}. %\footnote{The repository will be added if the article is accepted, for anonymization purposes.}. \todo{create a repo for this part of the project or update and make public the existing one}.
Experiments presented in this paper were carried out using computational clusters equipped with GPU from the Grid'5000 testbed (see \url{https://www.grid5000.fr}). Namely, the \textit{grele} and \textit{grue} clusters of were used. %\todo{Anonymize this}

\subsection{Dataset\label{sec:data}}
For our experiments, we use two datasets: the Sigmorphon2016~\cite{sigmorphon:2016:cotterell} dataset and the Japanese Bigger Analogy Test Set~\cite{jap-data:2018:karpinska}\footnote{\url{https://vecto.space/data/}}.

\subsubsection{Sigmorphon2016}
The Sigmorphon2016 dataset~\cite{sigmorphon:2016:cotterell} is a large inflexion dataset over 10 languages: Arabic (Romanized), Finnish, Georgian, German, Hungarian, Maltese, Navajo, Russian, Spanish, and Turkish. Most of those languages are considered to have rich inflexions.
It is separated in 3 subtasks: inflexion, reinflexion and unlabled reinflexion.
In our experiments, we focus on the data from the inflexion task, which is composed of triples of a source lemma $A$ (ex: ``do''), a set of features $F$ (ex: present participle) and the corresponding inflected form $B$ (ex: ``doing''). For example from the Finish data we have the triple: $A=$``lenkkitossut'', $F=$``pos=N,case=ON+ESS,num=PL'' (the transformation corresponds to the nominative to essive cases of a noun, for the plural) and $B=$``lenkkitossuilla''.

To obtain morphological analogies from this dataset, for any pair of triples $\langle A, F, B\rangle,\langle A', F', B'\rangle$ that have the same set of features ($F=F'$), we consider $A:B::A':B'$ an analogical proportion. Note that for each pair of triple, we only generate one analogy, \textit{i.e.}, if we generate $A:B::A':B'$ we do not generate $A':B'::A:B$ as it will be generated by the data augmentation process. Also, analogies of the form $A:B::A:B$ will be generated as the set of features is the same ($F=F$).
The number of analogies obtained by this process is detailed in \cref{tab:sigmorphon}.
During training and evaluation we generate for each analogy all the permutations and negative examples as mentioned in \cref{sec:data-augment}, resulting in a total of 8 valid analogies and repressively 3 and 24 invalid analogies for training and evaluation.

Several approaches were proposed to solve these subtasks, they can be grouped into three trends as mentioned in \cite{sigmorphon:2016:cotterell}: NLP pipelines using edit operations, neural approaches, and approaches based on linguistic heuristics.
None of the systems originally submitted relied on analogy solving, which is the approach we explore in this paper, in a similar manner as the approach to the task developed in \cite{minimal-complexity:2020:murena}.

\begin{table}[htbp]

        \centering
        \caption{Number of analogies for each language before data augmentation. We do not use the development set in our experiments.}
        \begin{tabular}{lrrr}\toprule
            \textbf{Language}   & \textbf{Train} & \textbf{Dev} & \textbf{Test} \\ \midrule
            Arabic              & 373240 & 7671 & 555312 \\
            Finnish             & 1342639 & 22837 & 4691453 \\
            Georgian            & 3553763 & 67457 & 8368323 \\
            German              & 994740 & 17222 & 1480256 \\
            Hungarian           & 3280891 & 70565 & 66195 \\
            Maltese             & 104883 & 3775 & 3707 \\
            Navajo              & 502637 & 33976 & 4843 \\
            Russian             & 1965533 & 32214 & 6421514 \\
            Spanish             & 1425838 & 25590 & 4794504 \\
            Turkish             & 606873 & 11518 & 11360 \\\bottomrule
        \end{tabular}
        \label{tab:sigmorphon}
    \end{table}

\subsubsection{Japanese Bigger Analogy Test Set}
The Japanese Bigger Analogy Test Set contains pairs of words related by a particular morphological relation, each relation is separated in a particular file.
From the available relations we focus on inflectional and derivational morphology, and process the data in a similar manner as Sigmorphon2016:
for every two pairs of words $\langle A, B\rangle,\langle A', B'\rangle$ related by the same relation we generated an analogical proportion $A:B::A':B'$, and we only generate one analogy per pair of word pairs. 
The selected relations correspond to 1017 pairs of words and 26410 analogies before augmentation.

    \begin{table}[htbp]
        \centering
        \caption{List of relations between the words of the Japanese dataset and corresponding number of unique analogies before data augmentation.}
        \begin{tabular}{clll}
        \toprule
        & \textbf{Relation} & \textbf{Example} & \textbf{Pairs}\hspace{-1em} \\
        \midrule
        \multirow{10}{*}{\STAB{\rotatebox[origin=c]{90}{Inflectional morphology}}}
            & verb\_dict - mizenkei01  & \begin{CJK}{UTF8}{min}会う $\rightarrow$ 会わ/あわ\end{CJK}  & 50 \\
            & verb\_dict - mizenkei02  & \begin{CJK}{UTF8}{min}出る $\rightarrow$ 出よ/でよ\end{CJK}  & 51 \\
            & verb\_dict - kateikei    & \begin{CJK}{UTF8}{min}会う $\rightarrow$ 会え/あえ\end{CJK}  & 57 \\
            & verb\_dict - teta        & \begin{CJK}{UTF8}{min}会う $\rightarrow$ 会っ/あっ\end{CJK}  & 50 \\
            & verb\_mizenkei01 - mizenkei02    & \begin{CJK}{UTF8}{min}会わ $\rightarrow$ 会お/あお\end{CJK}  & 50 \\
            & verb\_mizenkei02 - kateikei      & \begin{CJK}{UTF8}{min}会お $\rightarrow$ 会え/あえ\end{CJK}  & 57 \\
            & verb\_kateikei - teta    & \begin{CJK}{UTF8}{min}会え $\rightarrow$ 会っ/あっ\end{CJK}  & 50 \\
            & adj\_dict - renyokei     & \begin{CJK}{UTF8}{min}良い $\rightarrow$ 良く/よく\end{CJK}  & 50 \\
            & adj\_dict - teta         & \begin{CJK}{UTF8}{min}良い $\rightarrow$ 良かっ/よかっ\end{CJK}  & 50 \\
            & adj\_renyokei - teta     & \begin{CJK}{UTF8}{min}良く $\rightarrow$ 良かっ/よかっ\end{CJK}  & 50 \\
        \midrule
        \multirow{10}{*}{\STAB{\rotatebox[origin=c]{90}{Derivational morphology}}}
            & noun\_na\_adj + ka   & \begin{CJK}{UTF8}{min}強 $\rightarrow$ 強化/きょうか\end{CJK}  & 50 \\
            & adj + sa             & \begin{CJK}{UTF8}{min}良い $\rightarrow$ 良さ/よさ\end{CJK}  & 50 \\
            & noun + sha           & \begin{CJK}{UTF8}{min}筆 $\rightarrow$ 筆者/ひっしゃ\end{CJK}  & 50 \\
            & noun + kai           & \begin{CJK}{UTF8}{min}茶 $\rightarrow$ 茶会/ちゃかい\end{CJK}  & 50 \\
            & noun\_na\_adj + kan  & \begin{CJK}{UTF8}{min}同 $\rightarrow$ 同感/どうかん\end{CJK}  & 50 \\
            & noun\_na\_adj + sei  & \begin{CJK}{UTF8}{min}毒 $\rightarrow$ 毒性/どくせい\end{CJK}  & 52 \\
            & noun\_na\_adj + ryoku  & \begin{CJK}{UTF8}{min}馬 $\rightarrow$ 馬力/ばりき\end{CJK}  & 50 \\
            & fu + noun\_reg       & \begin{CJK}{UTF8}{min}利 $\rightarrow$ 不利/ふり\end{CJK}  & 50 \\
            & dai + noun\_na\_adj  & \begin{CJK}{UTF8}{min}事 $\rightarrow$ 大事/だいじ\end{CJK}  & 50 \\
            & jidoshi - tadoshi    & \begin{CJK}{UTF8}{min}出る $\rightarrow$ 出す/だす\end{CJK}  & 50 \\
            \midrule
        &\multicolumn{2}{l}{Total} & 1017 \\
        \bottomrule
        \end{tabular}
        \label{tab:japanese-relations}
    \end{table}
    
\subsubsection{Data for Training and Testing}
The model trained on Japanese was trained on %17775 randomly selected analogies (70\%) and tested on the remaining 7618.
18487 randomly selected analogies (70\%) and tested on the remaining 7923.
On Sigmorphon2016, as the amount of available data is very large, the models of each language were trained on 50000 and tested on 50000 analogies randomly selected from the train and test data respectively. After data augmentation, this correspond to 400000 positive and 150000 negative examples for training, and 1200000 negative examples for testing. However, Maltese, Navajo and Turkish were tested on only 3707, 4843 and 11360 analogies respectively due to the smaller size of the test dataset for those languages.

To maintain a reasonable training time, we trained the models for 20 epochs which correspond to around 6h with our computational resources.

\subsubsection{Lepage's analogies from Sigmorphon2016}\label{sec:lepage-sigmorphon}
For some of our comparability experiments we use the analogies extracted from Sigmorphon2016~\cite{sigmorphon:2016:cotterell} by Lepage \cite{sigmorphon:2017:lepage}, as it is the dataset used to compare the baselines in \cite{minimal-complexity:2020:murena}.
This dataset contains analogies extracted from Sigmorphon2016 Task 1 training data by aligning features in a similar manner as what we do. However, for each analogy $A:B::C:D$, the dataset also contains the following permutations:
\begin{itemize}
    \item $D:B::C:A$;
    \item $D:C::B:A$;
    \item $C:A::D:B$.
\end{itemize}
We notice that not all the permutations we use (detailed in \cref{sec:data-augment}) are considered.
\cref{tab:coverage} presents the portion of the analogies appearing in our dataset that is also appearing in Lepage's dataset (that we call coverage by Lepage's dataset of ours), as well as the converse coverage by our dataset of Lepage's. For our computations, we considered that if any permutation (among the 8 mentioned in \cref{sec:data-augment}) of an analogy is present in the other dataset, then the dataset covers this analogy.
Surprisingly, neither version completely covers the other, which means that even though our version tends to cover most of Lepage's version, the latter somehow contains some analogies that were not extracted using our process. However, as not much information is provided in \cite{sigmorphon:2017:lepage} about the extraction process, we were not able to determine how our processes differ in, other than the data augmentation process which was circumvented in our coverage computations.

\begin{table}[htbp]
    \centering
    \caption{Coverage (in \%) between Lepage's version of Sigmorphon2016 and the training set of our version. The coverage of the test set is 0\% in both directions for all languages and was not included.}
    \begin{tabular}{lcc}
    \toprule
     & \textbf{Coverage by Lepage's} & \textbf{Coverage by our}\\
    \textbf{Language} & \textbf{version of ours} & \textbf{version of Lepage's}\\
    \midrule
Arabic & 26.73 & 60.26\\
Finnish & 21.51 & 92.30\\
Georgian & 68.22 & 78.51\\
German & 68.07 & 92.38\\
Hungarian & 33.14 & 37.39\\
Maltese & 16.70 & 61.82\\
Navajo & 7.09 & 10.18\\
Russian & 25.81 & 91.81\\
Spanish & 54.23 & 91.44\\
Turkish & 29.12 & 71.80\\
    \bottomrule
    \end{tabular}
    \label{tab:coverage}
\end{table}

\subsection{Baselines}\label{sec:baselines}
To compare the performance of our model, we use the methods presented in   \cite{minimal-complexity:2020:murena}, \cite{tools-analogy:2018:fam-lepage}, and \cite{analogy-alea:2009:langlais}.
However, as the models from \cite{minimal-complexity:2020:murena,analogy-alea:2009:langlais} are not designed for classification of quadruples into analogies and non-analogies, we perform several adaptations.

\subsubsection{Kolmogorov Complexity Based Approach}
We use the generation model proposed by Murena \textit{et al.}~\cite{minimal-complexity:2020:murena} as a first baseline.
As the minimal-complexity approach is designed as a generative model, to have a classification for a given analogy $A:B::C:D$ we generate the top $k$ most likely solutions for $A:B::C:x$, in increasing order of Kolmogorov complexity. If $D$ is within the solutions, we consider $A:B::C:D$ as a valid analogy, otherwise we consider it a non-analogy. 
We use multiple threshold values from $1$ (only the most likely answer) to $10$, resulting in multiple baselines.
As the accuracy for $1<k<10$ is strictly within the range of the accuracy for $k=1$ and $k=10$, only $k=1$ and $k=10$ are displayed in the results, that we refer to as \lstinline{Kolmo@1} and \lstinline{Kolmo@10}.

\subsubsection{Alea}
We use the \lstinline{Alea} algorithm~\cite{analogy-alea:2009:langlais} as a second baseline.
This approach relies on a random process to generate potential permutations. %\todo{expand explanation or introduce in related works}
 For each quadruple $A:B::C:D$, we generate $\rho=1000$ potential solutions for $A:B::C:x$ and select the $k$ most frequent ones.
This approach is a generation model like the approach based on Kolmogorov complexity, so we use a similar process of generating potential solution and searching the expected $D$ among them.
Similarly to with the Kolmogorov complexity based approach, we use multiple threshold values from $1$ (only the most likely answer) to $10$, and report only the results with $k=1$ and $k=10$, that we refer to as \lstinline{Alea@1} and \lstinline{Alea@10}.

\subsubsection{Lepage Classifier}
We use the analogy classifier (\lstinline{is_analogy} in \lstinline{nlg/Analogy/tests/nlg_benchmark.py}) from Fam and Lepage's toolset \cite{tools-analogy:2018:fam-lepage} to classify analogies in the same manner as with our deep learning model.
This baseline uses a matrix based approach to align characters between words and find common sub-words, and then determine whether there is an analogy.
We refer to this baseline as \lstinline{Lepage}.
%\todo[inline]{add description of baseline}

\subsection{Results}
    \begin{table*}[htbp]
        \centering
        \caption{Accuracy results (in \%) for the classification task for positive ($+$) and negative ($-$) examples, as well as examples before augmentation (Base) for the Kolmogorov complexity based approach. Italicized values were obtained over a smaller subset of the data due to the relative slowness of the approaches.}
        \begin{tabular}{l|cc|ccc|ccc|cc|cc|cc}\toprule
                                &\multicolumn{2}{c|}{\textbf{CNN}} & \multicolumn{3}{c|}{\textbf{Kolmo@1}} & \multicolumn{3}{c|}{\textbf{Kolmo@10}} & \multicolumn{2}{c|}{\textbf{Alea@1}} & \multicolumn{2}{c|}{\textbf{Alea@10}} & \multicolumn{2}{c}{\textbf{Lepage}} \\
            \textbf{Language}   & $+$ & $-$ &     Base&$+$ & $-$  &Base & $+$ & $-$& $+$ & $-$& $+$ & $-$& $+$ & $-$  \\\midrule
            Arabic      & \textbf{99.89} & 97.52 & 42.59&28.94&\textbf{97.79}& 45.73&33.55&97.68 &28.28&97.68& 34.21&97.68    &32.94   &97.67 \\ %no significant difference for neg
            Finnish     & \textbf{99.44} & 82.62 & \textit{28.45}&\textit{22.82}&\textit{98.12}&--&--&-- &\textit{21.83}   &\textit{\textbf{98.78}}& --&--  &25.60   &98.05\\
            Georgian    & \textbf{99.83} & 91.71 & 91.68&80.19&95.01& 92.41&93.20&94.61 &86.81&\textbf{95.21} &91.68&95.21 &89.78   &95.18\\ %
            German      & \textbf{99.48} & 89.01 & 85.79&55.61&96.84& 86.35&60.27&96.65 &85.46&\textbf{97.19} &86.90&97.18 &83.32   &96.86\\ %
            Hungarian   & \textbf{99.99} & \textbf{98.81} & 36.19&31.21&98.40& 37.79&36.80&98.23 &35.79&98.24  &36.58&98.24    &33.81   &98.25\\ %
            Maltese     & \textbf{99.96} & \textbf{77.83} & 77.99&68.84&69.29& 82.03&73.64&67.32 &74.49&67.35  &78.05&67.32    &73.63   &67.32\\
            Navajo      & \textbf{99.53} & 90.82 & 18.54&17.97&\textbf{94.93}& 24.34&21.45&94.45 &16.16&94.76  &18.34&94.76    &17.35   &94.76\\
            Russian     & \textbf{97.95} & 79.85 & 44.28&33.37&93.66& 45.70&36.43&93.30 &42.00&93.74  &42.37&93.72    &41.02   &\textbf{93.88}\\
            Spanish     & \textbf{99.94} & 78.33 & 83.62&73.86&86.59& 84.24&81.54&86.44 &85.23&86.13  &85.90&86.13    &83.25   &\textbf{86.62}\\
            Turkish     & \textbf{99.48} & \textbf{92.63} & 43.63&39.37&91.40& 46.83&43.51&90.78 &42.53&90.80  &44.76&90.80    &34.92   &90.80\\\midrule
            Japanese    & \textbf{99.99} & \textbf{98.65} &  3.82&18.62&98.13&  3.85&19.20&98.09 & 3.75&98.14  & 3.77&98.14    &3.74    &98.16\\
            \bottomrule
        \end{tabular}
        \label{tab:results}
        \centering
    \end{table*}
\begin{table}[htbp]
        \centering
        \caption{Accuracy (in \%) of our model on Lepage's analogies from Sigmorphon2016. The baseline results are taken from \cite{minimal-complexity:2020:murena}, as regression accuracy corresponds to classification accuracy when using a regression model to classify analogies.}
        \begin{tabular}{lcc}\toprule
            \textbf{Language} & \textbf{CNN} & \textbf{Best baseline according to \cite{minimal-complexity:2020:murena}}\\\midrule
            Arabic              & \textbf{98.75} & 93.33 (Lepage)\\
            Finnish             & 93.57 & \textbf{93.69} (Kolmo)\\
            Georgian            & \textbf{99.56} & 99.35 (Kolmo)\\
            German              & \textbf{99.56} & 98.84 (Kolmo)\\
            Hungarian           & \textbf{99.32} & 95.71 (Kolmo)\\
            Maltese             & \textbf{97.93} & 96.38 (Kolmo)\\
            Navajo              & \textbf{99.82} & 86.87 (Lepage)\\
            Russian             & \textbf{99.61} & 97.26 (Lepage)\\
            Spanish             & \textbf{97.37} & 96.73 (Kolmo)\\
            Turkish             & \textbf{99.77} & 89.45 (Kolmo)\\\bottomrule
        \end{tabular}
        \label{tab:sigmorphon-lepage}
    \end{table}
\begin{table}[htbp]
    \centering
    \caption{Particularities (mean $\pm$ std. (max)) of the words involved in the analogies.}
    \begin{tabular}{lcc}
    \toprule
    {} &  \textbf{Average number of} & \textbf{Average} \\
    \textbf{Language} &  \textbf{adjacent repeated letters}& \textbf{word length} \\
    \midrule
    Arabic    &                                                 0.280 $\pm$ 0.463 (2) &   8.22 $\pm$ 2.23 (20) \\
    Finnish   &                                                 1.036 $\pm$ 0.908 (6) &  11.01 $\pm$ 3.63 (44) \\
    Georgian  &                                                 0.022 $\pm$ 0.148 (2) &   8.11 $\pm$ 2.21 (21) \\
    German    &                                                 0.204 $\pm$ 0.432 (3) &  10.37 $\pm$ 3.47 (35) \\
    Hungarian &                                                 0.159 $\pm$ 0.384 (2) &   8.69 $\pm$ 3.03 (20) \\
    Maltese   &                                                 0.663 $\pm$ 0.646 (3) &   9.43 $\pm$ 3.55 (19) \\
    Navajo    &                                                 0.851 $\pm$ 0.695 (2) &   7.69 $\pm$ 2.10 (17) \\
    Russian   &                                                 0.080 $\pm$ 0.278 (2) &   8.81 $\pm$ 2.83 (22) \\
    Spanish   &                                                 0.090 $\pm$ 0.293 (2) &   8.94 $\pm$ 2.31 (33) \\
    Turkish   &                                                 0.044 $\pm$ 0.206 (1) &   8.80 $\pm$ 3.61 (25) \\
    \midrule
    Japanese  &                                                 0.029 $\pm$ 0.174 (2) &   4.99 $\pm$ 3.19 (18) \\
    \bottomrule
    \end{tabular}
    \label{tab:words}
\end{table}
The classification accuracy for the baselines and our neural classification model (that we refer to as \lstinline{CNN}) are presented in \cref{tab:results}.

The Kolmogorov complexity and Alea baselines are slower on some languages like Spanish and German than on other languages. In fact, both approaches are extremely slow on Finnish: while for other languages computations finish (for 50000 examples) in less than 24h, the Kolmogorov complexity approach processes about 4000 examples of the Finnish data in that timeframe whereas Alea is even slower (only 300 analogies processed). For those two baselines we did not compute with a threshold of $k=10$ on Finnish as the computation time was too large.
Further analysis of the specificity of the words involved in the analogies (see \cref{tab:words}) confirms that the total length of the manipulated words and the number of repeated adjacent letters are correlated with this reduced speed of the Kolmogorov complexity and Alea baselines. In particular, Finnish and Navajo have a high number of adjacent repeated letters, and both Finnish and German have considerably large words. However, while both German and Navajo are slower than other languages, only Finnish is extremely slow. Thus, it is very likely that both characteristics must be present in the words manipulated in order to entail long computation time.
%\todo{one or the other = slow; both at the same time = very slow}

For our three baselines, we notice a strong imbalance of performance between positive and negative examples.
Analysis of where the model fails also leads us to testing the baselines on the analogies before applying the augmentation process (see \cref{sec:data-augment}), with the results reported in the ``Base'' column of \cref{tab:results} for the Kolmogorov complexity approach. 
For the other approach the results were not significantly different (here and after, by significant we mean that the $p$-value of student's t test is lower than 0.05) from those of the positive examples, so we did not report them in \cref{tab:results}.
The results on those ``raw'' analogies are significantly higher than those on the positive examples for the Kolmogorov complexity baseline, which hints that this approach is not resilient to the permutations performed when augmenting data. In particular, it tends to fail when central permutation is involved.
For further comparability with the three baselines, we report in \cref{tab:sigmorphon-lepage} the results of our approach when tested on the analogies extracted by Lepage \cite{sigmorphon:2017:lepage} (see \cref{sec:lepage-sigmorphon}). In \cref{tab:sigmorphon-lepage} we also showcase the best accuracy results reported in \cite{minimal-complexity:2020:murena} together with the corresponding model. Although results were computed as ``the proportion of correct answers when solving analogies''~\cite{minimal-complexity:2020:murena}, this corresponds to accuracy results when using  regression models as classifiers. On all languages our model outperforms the baselines, except for Finnish where the performance is comparable.

Moreover, we could expect the performance for the raw positive examples (``Base'' column) to match the one reported in \cite{minimal-complexity:2020:murena}, even if both versions of the dataset do not overlap exactly (see \cref{sec:data}).
A closer look at the suggested answers shows that this difference is due to a different preprocessing applied to the Sigmorphon2016: while we use all the analogies we can build using our simple alignment of features, Murena \textit{et al.}~\cite{minimal-complexity:2020:murena} use a variant where all the analogies that cannot be inferred by relying on only the source terms have been removed (for example in Finnish, the correct answer to the analogy ``mäyrä:mäyräss\underline{ä}::kolo:koloss\underline{a}'' is dictated by the rules of vowel harmony).
This means that some of the analogies in our version of the dataset are impossible to solve without additional data nor knowledge of the language (neither by an algorithm nor by a human). 
While the symbolic approaches of \cite{tools-analogy:2018:fam-lepage,minimal-complexity:2020:murena,analogy-alea:2009:langlais} do not rely on the languages they manipulate, our approach is trained specifically for a language, which allows it to infer the underlying rules and to perform well even for those analogies.
The proportion of such analogies varies from one language to another (according to the selection of the analogy, but also to the rules of the language itself), but no precise measurement was done as it would most likely require a manual analysis of all the analogies.

Our model performs the best out of all the tested models for positive examples, even if the baselines achieve more than 90\% and 80\% on Georgian and Spanish, respectively.
For the negative examples (invalid analogies) there are three cases: \textit{(i)}~our model performs the best (Maltese, Turkish, Japanese), \textit{(ii)}~there is no significant difference between our model and the baselines (Arabic, Hungarian), or \textit{(iii)}~our model has a lower accuracy but in that case there is no significant difference between the baselines (Finnish, Georgian, German, Navajo, Russian, Spanish). %\todo{Compute signifiance, for now it's \textit{à vue de nez}}
For Russian and Finnish, our model has an accuracy lower than the baseline by at least 13\% for negative examples. However, those baselines are much worse on the positive examples (by at least a 58\%).

%Overall, we notice that while our model was exposed to 3 times more negative than positive examples during training, it performs worse on negative examples.
Overall, our model performs worse on negative than on positive examples, which is coherent with the training process, as we use less than half the number of negative than positive examples during training.
Additionally, while our model performs slightly worse than the baselines on most negative examples, it strongly outperforms them for the positive examples.

%Le coeur du problème est que vous avez travaillé avec un sigmorphon différent du nôtre. Notre jeu de données a été fourni par Lepage (disponible sur le lien avec son code) et a été filtré par rapport au jeu de données initial. Cela explique donc la différence de valeurs. 
%Maintenant, pourquoi cette différence est-elle si importante ? Les cas qui ont été filtrés correspondent à des analogies dont la réponse exacte ne peut pas être inférée avec uniquement la source. Typiquement, en finnois : "mäyrä : mäyrässä :: kaura : kaurassa". La bonne réponse dépend d'une règle de grammaire (l'harmonie vocalique) qui ne se visualise pas à partir d'un seul exemple. Ce sont donc des analogies impossibles à résoudre sans connaissance de domaine. Marie a repéré également des exemples pareils en arabe. Vous, en revanche, bénéficiez d'un apprentissage sur un large jeu de données, ce qui permet à votre réseau de neurones d'inférer ces règles qu'un seul exemple ne pourrait donner. 

%\subsubsection{Extra experiments (maybe we won't have it in the final version)}
%When testing with random quadruples $\langle A,B,C,D\rangle$ as extra negative examples, the CNN still performs well;
%However, when using $\langle A,B,C,D\rangle$ such that $A:B$ or $B:A$ and $C:D$ or $D:C$ appear in the dataset of valid analogies, the performance drop significantly even if the relations $A:B$ and $C:D$ are not the same. Extra training including this new kind of negative examples have been tried without much success yet.

\section{Towards a Transferable Neural Analogy Model?}\label{sec:transfer}

\begin{figure*}[htbp]
\centering
\subcaptionbox{Base}[.33\textwidth]{\includegraphics[width=.33\textwidth]{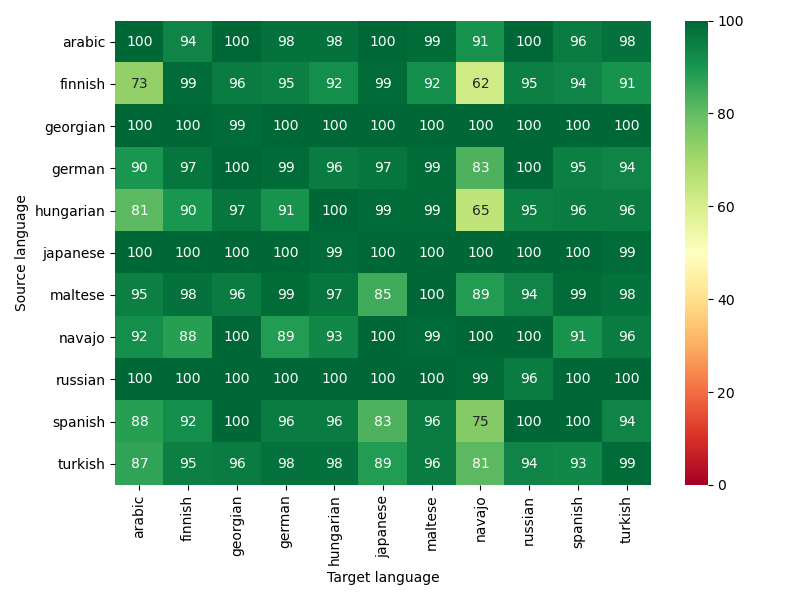}}%
\subcaptionbox{Positive}[.33\textwidth]{\includegraphics[width=.33\textwidth]{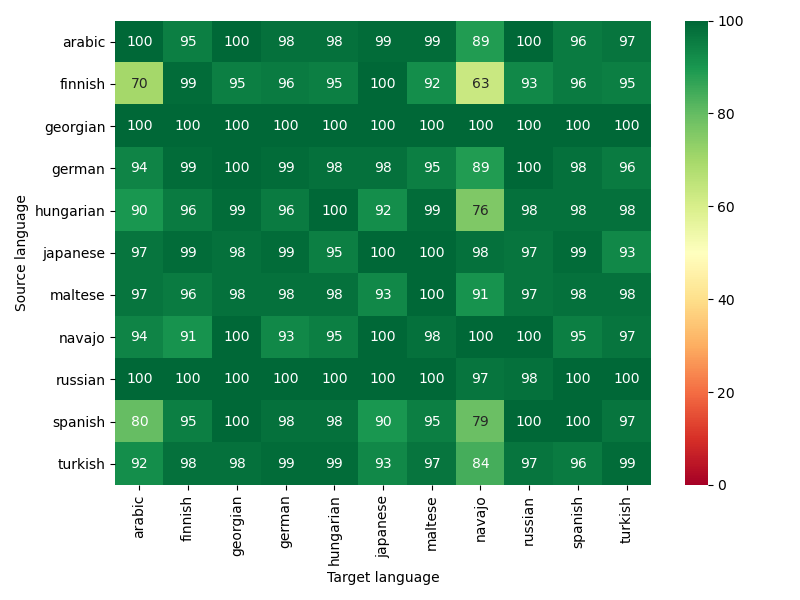}}%
\subcaptionbox{Negative}[.33\textwidth]{\includegraphics[width=.33\textwidth]{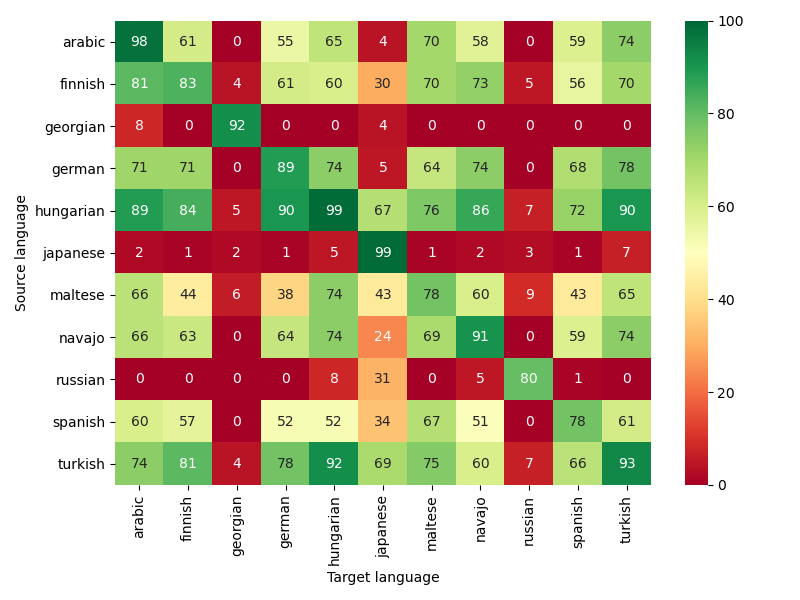}}
\caption{Accuracy (in \%) of fully transferred models (\textit{i.e.} embedding and classifier of a language applied to another language) on Sigmorphon2016, for each of the tree categories of examples.}
\label{fig:cmap-full}
\end{figure*}
\begin{figure*}[htbp]
\centering
\subcaptionbox{Base}[.33\textwidth]{\includegraphics[width=.33\textwidth]{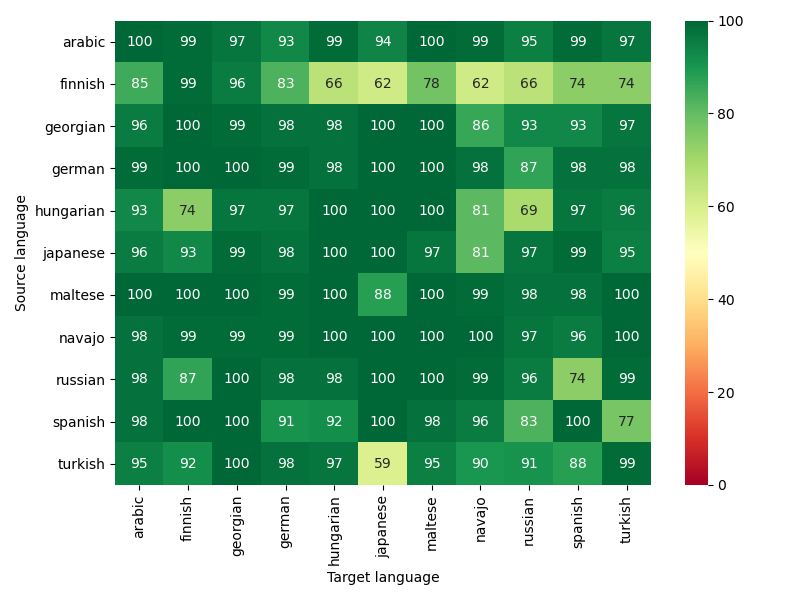}}%
\subcaptionbox{Positive}[.33\textwidth]{\includegraphics[width=.33\textwidth]{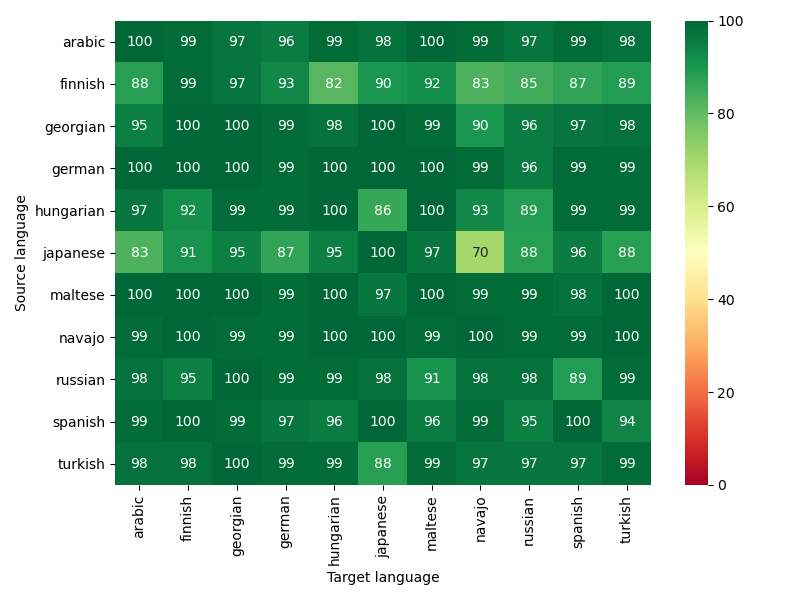}}%
\subcaptionbox{Negative}[.33\textwidth]{\includegraphics[width=.33\textwidth]{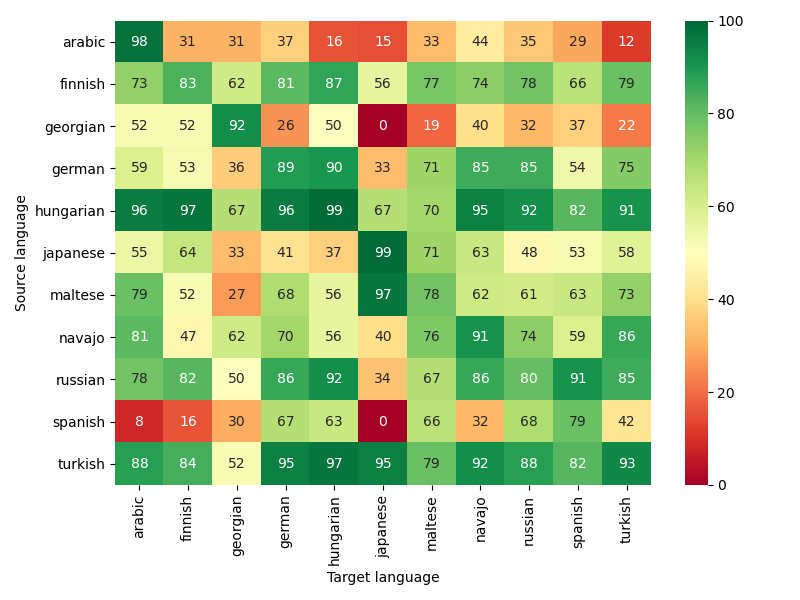}}
\caption{Accuracy (in \%) of partially transferred models (\textit{i.e.} classifier of a language applied to another language using the target language's embedding model) on Sigmorphon2016, for each of the tree categories of examples.}
\label{fig:cmap-part}
\end{figure*}

We applied the model trained on each language to the other languages of the dataset. The objective is to explore the generalization capabilities of the CNN model,  and to test its dependence on the training language.
In \cref{sec:transfer-all}, we present experiments where both the classifier and the embedding model were transferred from one language to an other.
We ran additional experiments using the embedding model trained on the target language and transferring only the classifier. These results are presented in \cref{sec:transfer-clf}.

\subsection{Embedding Model and Classifier of Another Language than the Target Language}\label{sec:transfer-all}
The results for full transfer (\textit{i.e.}, using both the embeddings and the classifier of a language on another language) are presented in \cref{fig:cmap-full}.

A quick comparison of the results shows that while our deep learning model performs well for base and positive examples, major differences appear for negative examples.
This is mostly due to differences in alphabet between the source and target language that cause a large portion of the target alphabet to be unrecognized.
This in turn leads to a large amount of 0 embeddings for the characters which tend to be ignored by the model.
%The results thus highlight that the classifier answers ``valid analogy'' by default, as the lack of incompatible features in the embeddings is in favor of a valid analogical proportion.\todo{make this sentence clearer}
%The high performance on positive and very low performance on negative classification may indicate that in case of unrecognized characters the classifier answers ``valid analogy'' by default. This makes sense as unrecognized characters are treated as padding by the word embedding model and thus should not impact the resulting embedding, and the embedding would be closer to that of an empty word (`` ''). As $A:B::C:D$ with $A=B=C=D=\text{`` ''}$ is a pretty obvious result, a direct consequence of not recognizing characters is the classification as a ``valid analogy''.
The high performance on positive and very low performance on negative classification may indicate that, in case of unrecognized characters, the classifier answers ``valid analogy'' by default. This makes sense as unrecognized characters are treated as padding by the word embedding model and thus should not impact the resulting embedding, and the embedding. The embedding would thus be closer to that of an empty word $\varepsilon$. As $\varepsilon:\varepsilon::\varepsilon:\varepsilon$ is obvious, a direct consequence of not recognizing characters is the classification as a ``valid analogy''.

Overall, most languages transfer reasonably well, with an accuracy between 50\% to 80\%.
However, models have an accuracy close to 0\% when transferred \textit{from} Georgian, Japanese and Russian or \textit{to} Georgian and Russian, which could be expected as most of the alphabet is not shared.
%Additionally, models transferred to Turkish and Hungarian perform slightly better compared to those transferred to other languages.
Additionally, models trained on Turkish and Hungarian perform slightly better when compared to those trained on other languages.
Interestingly, some models manage to transfer reasonably to Japanese despite the alphabet gap, in particular, from the above mentioned Hungarian and Turkish. 
Further analysis and experiments would be required to understand what happens in those cases.
Also, the quality of the transfer is not symmetric, which means that a model that performs well when transferred from a language~$A$ to a language~$B$ may perform badly when trained on language~$B$ and transferred to language~$A$.

\subsection{Embedding Model of the Target Language but Classifier of Another Language}\label{sec:transfer-clf}
The results for partial transfer (\textit{i.e.} using the classifier model of a language on another language using the target language's embedding model) are presented in \cref{fig:cmap-part}.

The results are similar to those of full transfer in that the model transfers well for base and positive examples while differences appear for negative examples.
However, we can notice a drop of up to 30\% in accuracy for base examples when transferring from Finnish.
For negative examples, the performance is above 10\% for all but three cases (from Georgian and Spanish to Japanese and from Spanish to Arabic), and above 25\% in most cases, which is a clear improvement from full transfer.
It is noteworthy that the target of the transfer seems less important than the source language, with Spanish, Georgian and Arabic, as well as Japanese to a lower degree, are languages from which transferred classifiers perform worse.
%This results matches the fact that embedding finetuned for a specific kind of data is performs much better than freshly transfered pretrained embeddings
%We can notice that transferring from Georgian and Japanese produces worse results that from other language.

This experiment display the performance of a model where the alphabet gap is not an issue, resulting in a significant improvement of the performance.
Nonetheless, the embedding model was not trained together with the classifier which may result in a mismatch in the representation.
Additional experiments are required to have a better understanding on what cause the word embedding space to mismatch that of the classifier.

\section{Conclusion}
We successfully adapted the approach of \cite{analogies-ml-perspective:2019:lim} from semantic to morphological analogies.
In this process, we discussed some of the limitations of the symbolic approaches with regard to analogy applied to morphology related tasks.

While the semantic approaches mentioned in this paper do not require training, our CNN model is more flexible as it is able to carry over domain and language specificities from the training process.
Our model is also more flexible in terms of the properties of analogies that it models. Changing the way the data is augmented will change the way the model behaves, which can allow the adaptation of the set of properties considered for analogies. For example, by dropping uniqueness or central permutation to fit another formalization of analogy. 
Additionally, our model has strong potential to carry over models of analogy when using an adapted embedding model as we showed with our transfer experiments. This stays true even if the embedding model does not perfectly fit the data. 

As we limited ourselves to analogy classification in this paper, further work includes testing the analogy solving aspect (the regression model) of Lim \textit{et al.}'s neural approach on morphology \cite{analogies-ml-perspective:2019:lim}.
Also, restricted ourselves to 3 instead of the 24 negative forms per raw analogy used in \cite{analogies-ml-perspective:2019:lim}.
Intuitively, using the full set of 24 forms for the negative examples during training will only increase the performance of our model, which would lead to outperform the state of the art for negative examples\footnote{This initial intuition has been attested, where better results were achieved for classification task when worked with 8 positive and 8 negative examples.}.
Furthermore, our early experiments on transferability highlight the potential to transfer and reuse our neural approach across domains.
The results also confirm our hypothesis that morphological analogy models are transferable between similar languages (in terms of alphabet and morphology). Further results on transferability could be used to measure morphological similarities between languages. 
All languages have their own characteristics in terms of the nature of used inflexions (suffixation, prefixation, duplication, etc). We intend to do a thorough study on the impact of these inflexions onto classifying (or solving) analogies, as well as onto the transferability between languages.

%The results of our transferability experiments 

%\todo{hypothèse: transférable entre langue proche, confirmée, pourait servervir à: The results could be compared to similarities between languages.
%Full training = even better}
\section*{Acknowledgments}
%For anonymization purposes, the Acknowledgments will be added if the article is accepted.
This research was partially supported by the project ``Foundations of Trustworthy AI integrating Learning, Optimisation and Reasoning'' (TAILOR) funded by EU Horizon 2020 research and innovation programme under GA No. 952215, and the Inria Project Lab ``Hybrid Approaches for Interpretable AI'' (HyAIAI).

Experiments  were carried out using the Grid'5000 testbed, supported by a scientific interest group hosted by Inria and including CNRS, RENATER and several Universities as well as other organizations (see \url{https://www.grid5000.fr}).

%\todo{MC: References need to be uniformised, compare e.g. 3, 7, 8. Also, others need to be reduced, e.g., 9.  }

%\vfill\eject
\bibliographystyle{IEEEtranN}
\bibliography{bib}

\end{document}